\documentstyle[11pt, a4]{article}

\if@twoside \oddsidemargin 44pt \evensidemargin 82pt \marginparwidth 107pt
\else \oddsidemargin -25pt \evensidemargin -25pt
\fi
\newcommand{\dson}{\renewcommand{\baselinestretch}{1.4}\large\normalsize}
\newcommand{\dsoff}{\renewcommand{\baselinestretch}{1}\large\normalsize}

\begin{document}
\begin{center}
\Large{\bf{The Grammar of Sense:}}\\
\large{Is word-sense tagging much more than part-of-speech
tagging?}\\
\vspace{0.15in}
\large{\bf{Yorick Wilks and Mark Stevenson\footnote{The authors are grateful to Mark Hepple, Robert Gaizauskas and Roberta
Catizone for many helpful comments and suggestions on this paper.}}}\\
\large{Department of Computer
  Science,\\University of Sheffield,\\Regent Court, 211 Portobello
  Street\\Sheffield S1 4DP, UK\\{\tt \{yorick, marks\}@dcs.shef.ac.uk}}\\
\end{center}
\vspace{0.15in}

\begin{abstract}
  
  This squib claims that Large-scale Automatic Sense Tagging of text
  (LAST) can be done at a high-level of accuracy and with far less
  complexity and computational effort than has been believed until
  now. Moreover, it can be done for all open class words, and not just
  carefully selected opposed pairs as in some recent work. We describe
  two experiments: one exploring the amount of information relevant to
  sense disambiguation which is contained in the part-of-speech field
  of entries in {\it Longman Dictionary of Contemporary English}
  (LDOCE). Another, more practical, experiment attempts sense
  disambiguation of all open class words in a text assigning LDOCE
  homographs as sense tags using only part-of-speech information. We
  report that 92\% of open class words can be successfully tagged in
  this way.  We plan to extend this work and to implement an improved
  large-scale tagger, a description of which is included here.

\end{abstract}

\vspace{0.15in}

\dson
\section*{Introduction}

Large-scale Automatic Sense Tagging (LAST) has become, like
part-of-speech tagging and automatic parsing before it, an important
and much researched intermediate task in natural language processing.
By ``intermediate'' we mean a task whose evaluation is determined by
some linguistic or theoretical criterion, as opposed to a task like
machine translation (MT) or information extraction (IE), where the
criteria can be judged by end-users of information, rather than those
familiar only with linguistic notations. This is an important distinction,
even if sometimes hard to make firm, but corresponds to a commonsense
intuition that no one wants part-of-speech tagging, parsing or LAST
information {\bf as such}, but only as a means to some other end,
unless, of course, the aim is to verify or refute a theory of language
processing or structure, where intermediate information is often
essential.

It is this difference of goals which has allowed intermediate tasks to
become so prominent in our field, sometimes at the expense of
``final'' tasks, since, as researchers, we are naturally more
preoccupied with the confirmation and refutation of theories than with
the provision of usable output. Depending on our particular taste in
theories, we tend therefore to elevate the status of certain
intermediate results because we believe, on a priori grounds, that
they are essential to the achieving of final goals. So, for example,
the importance of assessable large-scale corpus parsing has always
been more important to those who believe it to be an important step to
tasks like MT than to those who do not.

It has always been a clich\'{e}Ä of MT that the problem of word-sense
ambiguity was one of the intractable problems that has slowed progress
in achieving high quality MT, and that observation can be taken as
justifying attempts to achieve LAST. It remains to be seen whether
high achievement at any of these three intermediate tasks (now in
sight in all cases) does in fact raise the quality of MT and IE. (The
authors are currently working on a project which, in part, seeks to
discover how useful sense disambiguation will be for IE).

In the case of LAST, there is an additional problem about the
objectivity and assessability of the task, since the notion of
word-sense has proved more dubious and hard to make precise than that
of part-of-speech tag or syntactic category. In both those cases there
is broad agreement that, although there is a range of label sets
available, the sets are broadly mappable to each other, and that,
whatever the labels used, there is sufficient inter-subjective
agreement on syntactic structure. 

In the case of word-senses, there has not been such a consensus: it is
often observed that dictionaries may classify the senses of a given
word in different ways and that the sense classifications may be
incommensurable, in that there may be no mapping in terms of set
inclusion between the differing sense sets for a word in different
dictionaries. There is also the homograph problem: lexicographers
divide a word's usages into homographs and senses (where homographs
tend to be supersets of senses) in a way that resists clear analysis.
The normal explanation of homograph (that it is really a different
word that just happens to be spelled the same as another, in which
sense bank-for-money and bank-of-river are often deemed homographs in
English) does not allow one to decide, on etymological or
any other evidence, whether one is dealing with a homographic or sense
distinction.

Again, lexicographers are known to divide roughly between ``lumpers''
and ``splitters'': those who like to divide senses without apparent
end, and those who prefer larger (more ``homographic'') clusters of
usages. All this has led some to despair and call it all
sense-distinction, some going even so far as to say that words have
more or less one sense each (e.g. \cite{Waismann65}). Kilgarriff \cite{Kil93}
has argued that human subjects cannot in fact assign sense-tags to
words in corpora, which, if true would make LAST a pointless
enterprise. We have answered that claim elsewhere \cite{Wilks95}, but
it can leave others feeling that, at best, LAST can only produce a
tagging circular upon a particular dictionary, the one that contains
the sense tagging taxonomy used.

The response to that is to turn to the computational work that has
attempted to derive sense clusters directly from corpora without any a
priori ``seeding'' of the clusters. Work reported at IBM
\cite{Brown91} and by McDonald and Plate (see \cite{Wilks96}) using
quite different statistical techniques have shown that clusterings
correspond closely to a broad (close to homographic) notion of sense.
Again, one could cite the work of Itai and Dagon \cite{Dagon94} which
has shown the consistency of a cross-linguistic notion of word-sense
as determined by foreign language equivalents in bilingual texts, e.g.
the distribution of {\it ``duty''} in the Canadian Hansard English
texts against aligned French sentences that contain either {\it
  ``devoir''} (=duty as obligation) or {\it ``impot''} (=duty as tax),
where the broad distinctions are clear and obvious: the contexts of
{\it ``duty''} correspond to two objectively characterisable contexts
in French.

\subsection*{Work So Far}

All this leads us to believe that there is a proper intuition
underlying LAST, since those results are inter-subjective
beyond the framework of a particular sense taxonomy in a particular
dictionary. Existing LAST work cannot itself confirm that, but let us
review it quickly. Three basic methods have been used, corresponding to
intuitions of the linkage of word-sense to:
\begin{enumerate}
\item \label{item1} syntactic context, usually determined by the
  window of words in which a token occurs.
  
\item \label{item2} relevance to subject matter, in the sense of a
  topic context provided by, say, Roget's thesaurus heads (a method
  for LAST first explored by Masterman in 1966 (see \cite{Wilks96})).
  
\item \label{item3} overlap of word occurrences within the definitions
  of the senses to be distinguished, a method first proposed by Lesk
  \cite{Lesk86}.
\end{enumerate}

Method \ref{item3} has recently been optimised by Cowie and Guthrie
\cite{Guth92} using simulated annealing and they report results of
72\% correct assignment at the homographic level in LDOCE and a much
lower level for individual sense assignment. This result must be seen
against a background figure of 62\% \cite{Wilks96} correct sense
assignment in LDOCE achieved by assigning the {\bf first}\footnote{In
  LDOCE the senses are ordered by frequency of occurrence in text and
  so the first sense is the most likely.} LDOCE sense in an entry.
However, we suspect that the wrong optimisation function was used in
the annealing, one that tended to assign senses with long
definitions in LDOCE, and so the figure could have been much better, a
matter we intend to remedy later. The importance of this method is
that it disambiguates all the content words in a sentence, even though
it involved a vast computation for a sentence if all the LDOCE senses
were considered, often optimising more than $10^{9}$ sense
combinations for a 12 word sentence.

Yarowsky \cite{Yar93} has investigated both methods \ref{item1} and
\ref{item2}, and we have criticised his methods elsewhere, pioneering
though they are, and achieving figures of up to 96\% correct for
selected word distinctions \cite{Yar95}. The problems with his method
are that it is a very small scale method for numbers of words usually
less than 10. Moreover, although he has sought to combine methods, the
sense of ``sense'' used varies (from appearing under a single Roget
head to having a bilingual Dagon/Itai style-correspondence). One could
generalise and say his results can therefore be compared to Cowie and
Guthrie, though they are much superior on the smaller scale he uses,
since the distinctions Yarowsky makes (e.g. the Roget and bilingual
ones) are equivalent to what are distinguished as homographs in LDOCE.

A key fact to notice about \ref{item1}-\ref{item3} is that they are
methods resting on quite different intuitions: and one might well
infer that, if they all capture at least part of what we intuitively
mean by word-sense, then the sensible way to achieve high-quality
LAST is to combine all three (Yarowsky uses aspects of \ref{item1} and
\ref{item2}). In this squib, we shall show how we achieve
high percentage, large scale, figures with a method different from all
the above. In the conclusion we shall show how we intend to proceed by
combining our current results with aspects of \ref{item1},
\ref{item2}, and \ref{item3} to optimise our results further.

\section*{Starting again at LAST}

The observation with which we begin is that POS tagging and LAST are
not as independent as has always been assumed. POS tagging
\cite{Brill92} is a well established module in many NLP systems these
days giving accuracy figures of up to 98\%. Our first investigation
was to see how far, given a basic NLP lexicon such as a tractable form
of LDOCE, accurate POS tagging would discriminate senses without any
further processing. The result was far more striking than we
expected.\footnote{The authors are grateful to Mark Leisher at CRL in
  New Mexico State University who provided preliminary results,
  encouraging us to conduct further research.}


{\it The Longman Dictionary of Contemporary English \#1} \cite{Ldoce}
is a dictionary designed for students of English and containing around
36,000 word types. Each word type consists of one or more homographs,
the homographs themselves are sets of senses for the word type. Each
of the word-senses in LDOCE contains part-of-speech information
indicating which grammatical category that sense corresponds to, taken
from a set of 17 broad grammatical distinctions. All the senses which
make up a homograph have identical part-of-speech information.
However, this is not to say that word-senses are partitioned into
homographs by syntactic criteria: around 2\% of word types in LDOCE
contain a homograph which has more than one part-of-speech associated
with each of its senses, which is thus a homograph with multiple
parts-of-speech. There are also many words, for example {\it
  ``bank''}, which contain more than one homograph with the same
part-of-speech. We argue later that homographs partitioned by
grammatical categories are a natural side-effect of grouping
semantically related senses.

\subsection*{The Taxonomy of a Lexicon: A Gedankenexperiment}

We attempted to discover how useful part-of-speech information could
be for semantic disambiguation. We scanned through LDOCE and examined each
word type for possible disambiguation to the homograph level by part
of speech. By examining this information it is possible to place each
of the LDOCE word types in one of the following categories:

\begin{enumerate}  
\item {\bf Guaranteed disambiguation:} those word types for which no
  grammatical category is associated with more than one homograph.\\
These words will always be disambiguated if its part-of-speech is known.\\
eg. a word with 3 homographs with grammatical categories {\tt n}, {\tt
  v} and {\tt adj}. 
\item {\bf Possible disambiguation:} those for which there is at least one
  grammatical category associated with a single homograph but
  there is another category which is associated with more than one.\\
These words will be disambiguated by some part-of-speech assignments, but
  others will not disambiguate it fully.\\
eg. a word whose homographs had grammatical categories {\tt n}, {\tt
  v}, {\tt v}.
\item {\bf No disambiguation:} those for which each grammatical
  category that can apply to the word type is associated with more than one homograph.\\
  These word can never be fully disambiguated by part-of-speech alone.\\
  eg. homographs with grammatical categories {\tt v}, {\tt v}, {\tt
    n}, {\tt n}.
\end{enumerate}

The number of words which fall into the guaranteed disambiguation
category puts a lower bound on the number which a perfect
part-of-speech tagger could disambiguate, while an upper bound can
be found by adding the number which fall into the possible
disambiguation category, since these may be disambiguated by the
information contained in a part-of-speech tag, although they may not
be.

We examined each of the word types in LDOCE (except for closed class
words such as prepositions) and found that 34\% were polysemous and
12\% polyhomographic (a word types must, of course, be polysemous to
be polyhomographic since each homograph is a non-empty set of senses).
88\% of the polyhomographic words were guaranteed to be disambiguated
to the homograph level and 95\% of them could possibly be
disambiguated to the homograph level.  If we assume that all
monohomographic words are trivially disambiguated then we can translate
these values to 98.5\% guaranteed disambiguation and 99.4\% possible
disambiguation over all word types.


This experiment of course presumes a perfect POS tagger but, as we
have already mentioned, many fairly reliable taggers are readily
available. It is impossible to tell how these results will translate to a
real experiment since the results of this will be highly dependent
upon the distribution of word types across tokens in the corpus which is being
examined. In this Gedankenexperiment each of the word types in the dictionary
is considered only once, but some word types will occur many times in a
corpus and even more never will. So, for example, the upper bound
would not apply if, by chance, none of the words of type 3 appeared in
a given corpus.

\subsection*{Using a Tagger: An Aktionexperiment}

To test this method in practice we took five articles from
the Wall Street Journal, containing around 1700 words in total, and
disambiguated the content words using part-of-speech tags as the sole
information source.

The texts were POS tagged using the Brill tagger \cite{Brill92} and
open class words were flagged (the POS tags were used to decide which
words belonged to the open classes). The tags set used by the Brill
tagger were manually mapped onto the simpler part-of-speech fields for
LDOCE homographs.\footnote{The Brill tagger uses the tag set from the
  Penn Tree Bank which contains 48 tags \cite{ptb}, LDOCE uses a set
  of 17 more general tags.} The LDOCE homographs which corresponded to
the part-of-speech assigned by the tagger were extracted from the
appropriate LDOCE entry and the first (most frequent) of those was
chosen as the sense of the word.

We found that 92\% of the content word tokens were tagged with the
correct homograph compared with manual tagging of the same five texts.
57\% of the open class words were in fact polyhomographic and of these
87.4\% were assigned the correct homograph. The monohomographic words,
which made up the rest of the open class words were, trivially, 100\%
correct.

It is perhaps worth noting in passing that although only 12\% of the
word types in LDOCE are polyhomographic, more than half the content
words in an actual text are. This is an indication of the kinds of
words which are commonly used in English and is in keeping with Zipf's
Law \cite{zipf35}. 

Our simple and cheap method achieves a much higher result than the
computationally intensive method of Cowie and Guthrie with identical
coverage (all open class words in a text) and similar results to
Yarowsky's method with far greater coverage.

\section*{Conclusion}


Our result should not be misinterpreted as implying some kind of
reduction of semantic matters to syntactic or morphological ones and
so to a loss of richness of texture in NLP.  First, because
grammatical categories are themselves essentially semantic in origin,
a fact not contradicted by observing that many languages have
inflectional criteria for what it is to be a particular
part-of-speech. It is no answer to the question {\itÓ``What is a noun in
  German?''}Ô to answer that it is the part-of-speech that is regularly
capitalised!

The commonsense view is that parts-of-speech are rooted in our basic
ontology of the world, of how it is, which is in turn a fundamentally
semantic matter. In the philosophy of language this view is sometimes
thought weakened by observations like those of Waismann \cite{Waismann65} that
some aspects of the world are captured in one language by the use of
one part of speech in one language but by a different part-of-speech
in another, which, if true, implies the matter cannot be semantic in
the sense of how the world is independent of ourselves and our
languages. But this, fortunately for us as NLPers, is a question on
which we do not have to have views: it is certainly not an issue that
can divide parts-of-speech from word-sense as one that separates
language from the world, or at least the perceived world.

A more persistent worry this result may exacerbate is the
traditional AI view of all these matters, one shared with Bar Hillel,
that issues of word-sense were to be settled by world knowledge,
not again in any objective sense, but as a function of stored codings
that express the state of the world. If that view is right (and many
of the authors of say \cite{Small88} held it in 1988) then, it is
unacceptable that a crucial issue like word-sense be settled by
matters independent of stored world knowledge.

Matters are not really so depressing, and one way to construe the
current result is that low level methods can give a very effective,
basic notion of word-sense discrimination, probably close to what we
are calling a homograph, and that all finer distinctions, whether one
wishes to call them word-sense or not, are matters for world
knowledge, which is to say, classic AI. So, one can cite of simple
examples such as Ó{\it ``He wiped the bicycle before sitting on
  it.''}Ô which have been used to argue that there is a sense of
ÓbicycleÔ meaning {\it ``bicycle seat''}, and then so on for each of
its 250 component parts. This is plainly absurd: an extension of
word-sense into an area best thought of as knowledge processing.

The current techniques can be seen as defining (at least when
optimised, see below) a limit to the extension of Óword-senseÔ and
thus the demarcating the fields of NLP and AI-proper. This is perhaps
no more than a sensible compromise position, consistent with the NLP
discrimination methods available.

In conclusion, it is important to stress that our method is {\bf
  comparable} to other recent work like Yarowsky's and Cowie \&
Guthrie: like theirs our method separates senses with respect to an
available human-constructed database of substantial size (LDOCE in our
case, Roget, Groliers or bilingual text in his) and at about the same
level of grain size, namely the LDOCE homograph.

\section*{Further work}

What we plan now, and which should have produced results before this
squib appears, is to pipeline a number of independent sources of
sense tagging information together, probably in the following order:

\begin{enumerate}
\item \label{re2}LDOCE sense discrimination by POS
\item \label{re3}Subject codes (= LDOCE pragmatic codes)
\item \label{re4}LDOCE examples as correlates
\item \label{re5}Simulated annealing optimisation of Lesk's 
heuristic.
\end{enumerate}

\ref{re2} was described here. \ref{re3} has been used
 to produce a sense-tagged hierarchy for all LDOCE
nouns at a high level of accuracy \cite{Wilks96}, and is essentially the same type of
information as the Roget Thesaurus used by Yarowsky in
\cite{Yar95}. \ref{re4} is a limited version of the
One-sense-per-collocation heuristic of Yarowsky \cite{Yar93} which he
showed had sense resolving power for almost any explicit form of
collocation. We propose to use the example sentences in
LDOCE (or perhaps the Longman Activator \cite{activator}) for each sense as a possible signature correlate. These examples
have been much criticised, not being corpus attested, but they are
easily available on a large scale and, even if they prove a weak source
of information, will be unlikely to harm the sense filtering.
\ref{re5} is the Cowie and Guthrie method but with a changed
 optimisation function as we noted earlier. We have been developing GATE, a
General Architecture for Text Engineering \cite{gate}, and expect that
to provide the appropriate environment for developing this
 comprehensive multi-source word-sense discriminator.

\dsoff

\bibliographystyle{plain}

\end{document}